# Execution Semantics of Behavior Trees in Robotics Applications

Enrico Ghiorzi* UNIGE, IIT
Christian Henkel† BOSCH
Michaela Klauck‡ BOSCH
Matteo Palmas§ BOSCH, UNIGE
Alberto Tacchella¶ UNITN
Armando Tacchella‖ UNIGE

April 10, 2025

Behavior Trees (BTs) have found a widespread adoption in robotics due to appealing features, their ease of use as a conceptual model of control policies and the availability of software tooling for BT-based design of control software. However, BTs don't have formal execution semantics and, furthermore, subtle differences among implementations can make the same model behave differently depending on the underlying software. This paper aims at defining the execution semantics of behavior trees (BTs) as used in robotics applications. To this purpose, we present an abstract data type that formalizes the structure and execution of BTs. While our formalization is inspired by existing contributions in the scientific literature and state-of-the art implementations, we strive to provide an unambiguous treatment of most features that find incomplete or inconsistent treatment across other works.

*enrico.ghiorzi@edu.unige.it
†christian.henkel2@de.bosch.com
‡michaela.klauck@de.bosch.com
§matteo.palmas@de.bosch.com
¶alberto.tacchella@unitn.it
‖armando.tacchella@unige.it

# Introduction

In recent years, behavior trees (BTs) have found increasing applications in deliberation systems in robotics [5, 11, 4, 14, 9, 13]. Because of their growing popularity, a number of frameworks are available implementing BTs for robotic control systems [1, 2, 15, 16, 19]. Unfortunately, in spite of some proposals existing in the literature [6, 17, 12, 7, 18, 10, 3], BTs lack a reference formal semantics that could provide the basis for their implementation *and* their verification both when considered alone or in some embedding context, i.e., when BTs orchestrate functional components in a robot control architecture. Non-uniform treatment of various aspects of BT execution is common across different libraries, or even across different versions of the same library, and some concepts that are strongly required in robotics, e.g., the capability of halting the execution of a (sub)tree, are not well specified and their definition is left to implementations.

This paper is an attempt to give a precise and unambiguous definition of the basic concepts of BTs and the functioning of specific nodes by distilling from [5] and other authoritative reference including implementations like the BT.cpp library [1]. Our formalization is given in the form of code whose syntax is borrowed from the Java language and whose semantics can be formally defined, e.g., by referencing the standard semantics of the Java language itself, or by providing ad-hoc rules. Our ultimate goal is to provide a definition of BTs which is precise enough, yet easy to grasp also by non-experts in formal logic and that includes most of the features popularized in robotics by the literature and implementations that we are aware of.

The paper is organized as follows. In Section 1 we provide an informal introduction to BTs and their execution. Section 2 formalizes the structure and the content of BTs as an abstract data type: each kind node is defined as a class which specifies the attributes and methods available for each instance of that kind. In Section 3, we define how a BT is built using the previous definitions and we present a graphical syntax to describe such BTs, along with some examples. Finally, in Section 4 we discuss some of the open questions left for future work.

# 1 Behavior Trees

Informally, a *behavior tree* (BT) is an ordered tree having *control flow* and *decorator* nodes as internal nodes, and *execution* nodes as leaves [5]. A periodic signal called *tick* is sent to the root of the BT which will then emit a *response* within the tick period. The response is either *success*, *failure*, or *running* for a successful, unsuccessful or undetermined execution, respectively. The execution flow is the following:

- The root node is ticked.
- Control flow nodes activate when ticked and can propagate the tick to one of their children based on their control logic; children may be either control flow nodes or execution nodes.

- Execution nodes activate when ticked and respond `SUCCESS`, `FAILURE`, or `RUNNING`.
- Control flow nodes receive responses from their children and decide what to respond to their parents.
- Finally, by recursively back-propagating along the tree structure, a response reaches the root node.

To handle scenarios in which the execution of the BT needs to be stopped, we introduce the *halt* signal — available also in implementations like [1]. Informally, the *halt* signal works as follows:

- Execution nodes can be halted; it is their responsibility to propagate the signal to underlying functional elements and coordinate them to obtain the expected result — e.g., halting an action should cause the robot to stop performing that action.
- Control nodes can be halted and thereby propagate the signal to their children according to their control logic.
- Ticking and halting are blocking operations and the execution flow of the BT is suspended until the ticked or halted node returns.
- We assume that halting can happen quickly enough to avoid stalling the execution of the BT, i.e., the BT will respond to a halt signal within the tick period.

# 2 An Abstract Data Type for BTs

We formalize the various kinds of BT nodes as objects instantiated from the classes shown in Figure 1 — a combination of the *Composite*, *Decorator* and *Interpreter* design patterns [8]. The base class `Node` is presented in Listing 1 together with the definition of the enumeration `Response` which is the return type of the `tick` method, and the enumeration `State` which is the type of attribute `state` and the return type of the `getState()` method. `Node` has two attributes: `state` of type `State` is the internal state of the node as returned by the `getState()` method; `blackBoard` of type `SymbolTable` is a reference to a key-value data structure that can be used to extract and store data to configure the actual behavior of the nodes, e.g., by providing parameters that should be passed to the functional components that the BT orchestrates. Notice that `getState()` is concrete and its implementation is shared by all the subclasses so that all the nodes have a `state` which can be queried, but only the subclasses of `Node` are allowed to change it. The same goes for `blackboard`: although in principle every node can have its own reference to a specific symbol table, most implementations will refer to a single context provided by a unique data structure referenced by all the nodes. There are four different kinds of nodes corresponding to four abstract subclasses of `Node` detailed as follows:

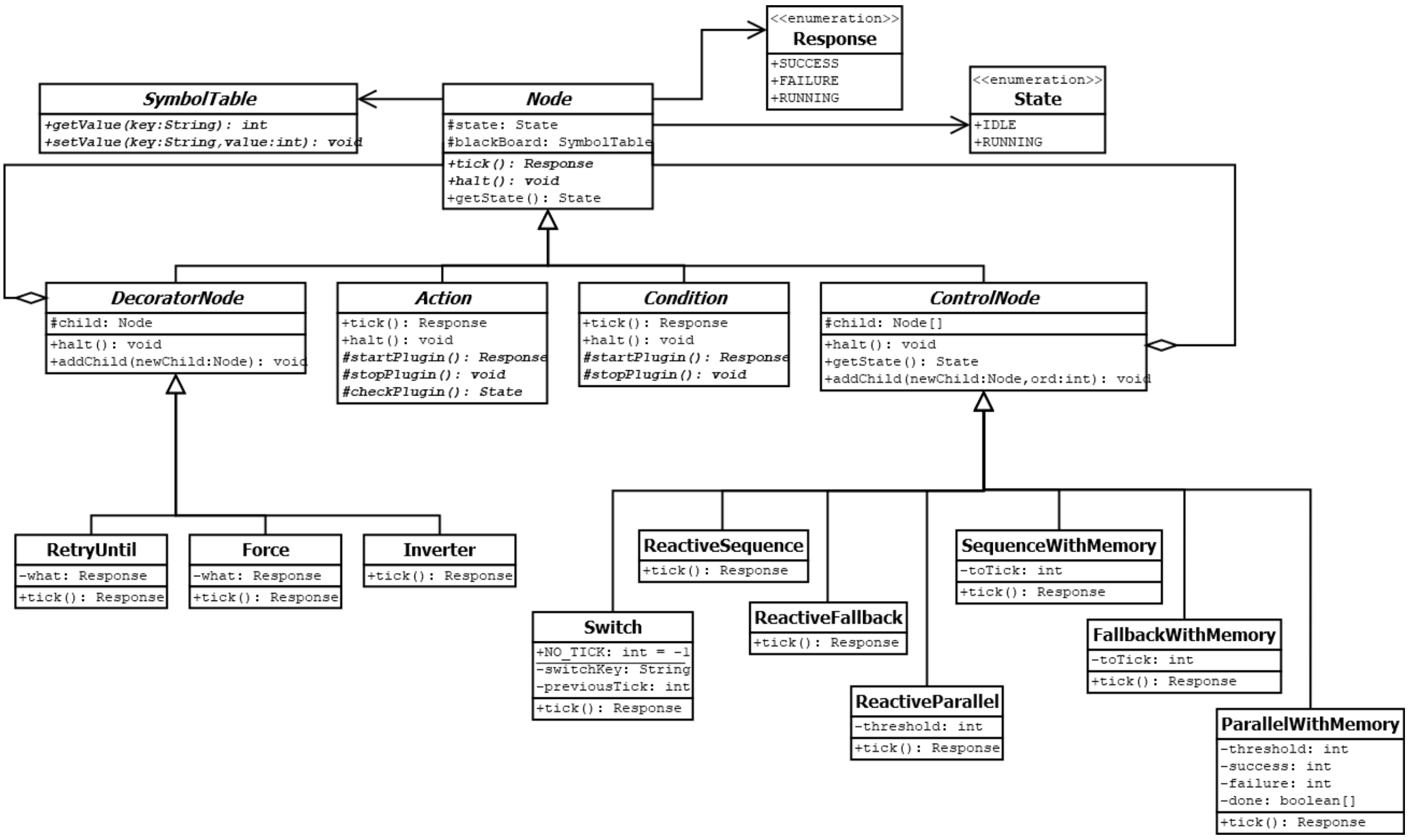

Figure 1: Class diagrams of BT nodes.

```java
enum Response {SUCCESS, FAILURE, RUNNING}
enum State {IDLE, RUNNING}

abstract class Node {

  protected State state;
  protected SymbolTable blackboard;

  public Node(SymbolTable blackboard) {
    this.state = State.IDLE;
    this.blackboard = blackboard;
  }

  public abstract Response tick();
  public abstract void halt();

  public State getState() {
    return this.state;
  }

}
```

Listing 1: `Response` and `State` enumerated types and abstract class `Node`.

| Colledanchise and Ögren [5] | *BehaviorTree.CPP* [1] | This document |
|---|---|---|
| Sequence | `ReactiveSequence` | `ReactiveSequence` |
| Sequence with memory | `Sequence` | `SequenceWithMemory` |
| *N.A.* | `SequenceWithMemory` | *N.A.* |
| Fallback | `ReactiveFallback` | `ReactiveFallback` |
| Fallback with memory | `Fallback` | `FallbackWithMemory` |
| *N.A.* | `FallbackWithMemory` | *N.A.* |
| Parallel | *N.A.* | `Reactive Parallel` |
| *N.A.* | `Parallel` | `ParallelWithMemory` |

Table 1: Nomenclature comparison of BT nodes among different references.

- Execution nodes `Action` and `Condition`. Both classes nodes override the base class methods with concrete implementations whose actual content depends on the *execution context* of the BT, i.e., the functional components that the BT orchestrates, and the protocols that are used to communicate with such components. The two classes differ on the constraints placed on the return values of `tick` and `halt` as detailed in subsection 2.1. Both classes are structured according to a *template method* design pattern: the skeleton of `tick` and `halt` is defined, but the actual interface with the embedding context is left to derived classes which should provide implementations for the (protected) abstract methods `startPlugin`, `stopPlugin` and `checkPlugin`.

- `ControlFlowNode` has several concrete subclasses. A group of six subclasses arises from the combination of three different control logics, namely *sequence*, *fallback* and *parallel* and two different ways of handling subsequent ticks, namely a purely *reactive* one and one endowed *with memory*. The `Switch` subclass is the implementation of a selection among children. In subsection 2.2, for each concrete subclass, we provide details of the overridden implementations for `tick()` and `halt()` methods.

- `DecoratorNode` has also several concrete subclasses (`Retry`, `Until`, `Force` and `Inverter`) whose details are presented in subsection 2.3.

```java
abstract class Action extends Node {

  protected abstract Response startPlugin();
  protected abstract void stopPlugin();
  protected abstract State checkPlugin();

  public Action(SymbolTable blackboard) {
    super(blackboard);
  }

  final public Response tick() {
    if (this.checkPlugin() == State.IDLE) {
      Response response = this.startPlugin();
      if (response == Response.RUNNING) {
        this.state = State.RUNNING;
        return Response.RUNNING;
      } else {
        this.stopPlugin();
        this.state = State.IDLE;
        return response;
      }
    } else {
      return Response.RUNNING;
    }
  }

  final public void halt() {
    if (this.checkPlugin() != State.IDLE) {
      this.stopPlugin();
    }
    this.state = State.IDLE;
    return;
  }

}
```

Listing 2: Abstract class `Action`.

## 2.1 Execution Nodes

The definition of `Action` abstract class is presented in Listing 2. The class provides a concrete implementation for the methods `tick` and `halt` which is not meant to be overridden by derived classes. However, derived classes should provide the implementation of three methods instead:

- `startPlugin` is meant to start whatever functional component the concrete action node interfaces to, e.g., a navigation module that makes a robot go from one place to another. As such, the return value of `startPlugin` can be any of `SUCCESS`, `FAILURE` or `RUNNING`. In the former two cases, it is assumed that the action is completed (un)successfully, whereas in the latter case it is assumed that the action is still being completed. We need this to implement the `halt()` semantics.

- `stopPlugin` is meant to stop any functional component that was previously started. It is assumed that this method does not return until stopping is complete.

- `checkPlugin` is meant to return the "state" of the underlying functional component, whether it was started and running, or sitting idle waiting for input.

Given the above methods, then a concrete subclass of `Action` may return `RUNNING`, `FAILURE` or `SUCCESS` when `tick()` is called, depending on the return values of `checkPlugin` and `startPlugin`. The value of the `state` attribute for an action node can be either `IDLE` or `RUNNING`: initially, `status` is `IDLE` — see the `Node` constructor in Listing 1; when `tick()` returns `SUCCESS` or `FAILURE`, `status` is assigned to `IDLE`; when `tick()` returns `RUNNING`, then `state` is assigned to `RUNNING`. When `halt()` is called, then `status` is always set to `IDLE` before the call returns, after stopping the plugin if its state is found to be `RUNNING`.

```java
abstract class Condition extends Node {

  protected abstract Response startPlugin();

  public Condition(SymbolTable blackboard) {
    super(blackboard);
  }

  final public Response tick() {
    Response response = this.startPlugin();
    assert (response == Response.SUCCESS ||
            response == Response.FAILURE);
    return response;
  }

  final public void halt() {
    return;
  }

}
```

Listing 3: Abstract class `Condition`.

The definition of the abstract class `Condition` is presented in Listing 3. Notice that this class has only the abstract method `startPlugin`: while classes deriving from `Action` may correspond to durative actions, the classes deriving from `Condition` are meant to be checks that should always require a negligible time (with respect to the whole tick time) to be performed. For this reason, the return value of `startPlugin` is constrained to be either `SUCCESS` or `FAILURE`. The state of a condition node is always `IDLE` and thus calling `halt()` has no effect.

```
abstract class ControlNode extends Node {

  protected Node[] child;

  public ControlNode(SymbolTable blackboard, int nChildren) {
    super(blackboard);
    assert(nChildren > 1);
    this.child = new Node[nChildren];
  }

  public void halt() {
    for (Node c : this.child) {
      if (c.getState() != State.IDLE) c.halt();
    }
    this.state = State.IDLE;
    return;
  }

  public void addChild(Node newChild, int ord) {
    assert(newChild != null);
    assert(ord>=0 && ord<child.length);
    this.child[ord] = newChild;
    return;
  }

}
```

Listing 4: Abstract class `ControlNode`.

## 2.2 Control Flow Nodes

We consider three different basic control logics that we can describe informally as follows:

- A *sequence* is meant to tick all of its children one after the other; if at least one child fails, the whole sequence fails; as long as at least one child is running, the whole sequence is running.
- A *fallback* is always going to tick the first child: if the response is successful, the whole fallback is successful and the other children are not ticked; if the first child fails, then the second one is ticked and so on, until either one child returns success, or all children fail; in the latter case, the whole fallback fails; as long as at least one child is running, the whole fallback is running.
- A *parallel* differs from a sequence in that subsequent children can be ticked even if they are still running; also, for a parallel node the overall success can be linked to the success of $k$ out of $n$ nodes with $1 \leq k \leq n$.

For each control logic we consider two variants: a *reactive* (i.e., memoryless) one, which does not keep track of the child that needs to be ticked next between subsequent ticks of the father, and one endowed *with memory* (i.e., memoryfull) that retains such information.

Before delving into the details of each single node, we must mention that the nomenclature in the literature is inconsistent and might not correspond to the one defined in this document. In Table 1 we present a comparison between this document, and the references [5] and [1]. The rationale behind our names is that each control flow node in its "purest" form is of the reactive kind, whereas any information retained between subsequent ticks is to be implemented by memoryful variants. We believe our nomenclature is in line with [5], once we take explicitly into account our rationale, and it is more consistent than the one given in [1], particularly when it comes to parallel nodes.

In Listing 4 we present the abstract class `ControlNode`. Control nodes are composites with at least two children — notice the `assert` directive that prevents the creation of control nodes with `nChildren` $< 2$. Control nodes do not override the definition of `getState()` but it is their responsibility to set the state properly when `tick()` is called. The definition of `halt()` will halt all the children which are not already `IDLE` and then setting the state of the control node to `IDLE`. Finally, it is possible to add a child in a specific order, as long as the new child is non `null` and the order is within the allocated range of the `child` array. The class remains abstract as it does not override the abstract method `tick` whose logic differs among the different concrete subclasses of `ControlNode`.

```java
final class ReactiveSequence extends ControlNode {

  public ReactiveSequence(SymbolTable blackboard, int nChildren) {
    super(blackboard, nChildren);
  }

  public Response tick() {
    for (int i = 0; i < this.child.length; ++i) {
      Response childResponse = this.child[i].tick();
      if (childResponse == Response.RUNNING ||
      childResponse == Response.FAILURE) {
        for (int j = i+1; j < this.child.length; ++j) {
          if (this.child[j].getState() != State.IDLE) {
            this.child[j].halt();
          }
        }
        if (childResponse == Response.RUNNING) {
          this.state = State.RUNNING;
        } else {
          this.state = State.IDLE;
        }
        return childResponse;
      }
    }
    this.state = State.IDLE;
    return Response.SUCCESS;
  }

}
```

Listing 5: Concrete class `ReactiveSequence`.

**Reactive Sequence** The `ReactiveSequence` concrete class overrides the `tick()` method as described in Listing 5. The tick is propagated to the all the nodes in the `child` array, from 0 to `child.length - 1`; if a child returns either `FAILURE` or `RUNNING` then all the remaining children which are not idle are halted and `FAILURE` or `RUNNING` is returned accordingly; if all children return `SUCCESS`, then `SUCCESS` is returned. Notice that when a child returns `RUNNING` or `FAILURE`, the tick is not propagated to the next child (if any). Also, the state of a reactive sequence is `RUNNING` only if at least one child is running.

```java
final class ReactiveFallback extends ControlNode {

  public ReactiveFallback(SymbolTable blackboard, int nChildren) {
    super(blackboard, nChildren);
  }

  public Response tick() {
    for (int i = 0; i < this.child.length; ++i) {
      Response childResponse = this.child[i].tick();
      if (childResponse == Response.RUNNING ||
          childResponse == Response.SUCCESS) {
        for (int j = i+1; j < this.child.length; ++j) {
          if (this.child[j].getState() != State.IDLE) {
            this.child[j].halt();
          }
        }
        if (childResponse == Response.RUNNING) {
          this.state = State.RUNNING;
        } else {
          this.state = State.IDLE;
        }
        return childResponse;
      }
    }
    this.state = State.IDLE;
    return Response.FAILURE;
  }

}
```

Listing 6: Concrete class `ReactiveFallback`.

**Reactive Fallback** The `ReactiveFallback` class redefines the `tick()` method as shown in Listing 6. The implementation of `tick()` is the "dual" of the same method in `ReactiveSequence`: the tick is propagated to the all the nodes in the `child` array, from 0 to `child.length - 1`; if a child returns either `SUCCESS` or `RUNNING` then all the remaining children which are not idle are halted and `SUCCESS` or `RUNNING` is returned accordingly; if all children return `FAILURE`, then `FAILURE` is returned. Also in this case, when a child returns `RUNNING` or `SUCCESS`, the tick is not propagated to the next child (if any) and the node is running only if at least one child is also running.

```java
final class ReactiveParallel extends ControlNode {

  private int threshold;

  public ReactiveParallel(SymbolTable blackboard, int nChildren,
                          int threshold) {
    super(blackboard, nChildren);
    assert(threshold>=1 && threshold<=nChilren);
    this.threshold = threshold;
  }

  public Response tick() {
    int success = 0;
    int failure = 0;
    for (int i = 0; i < this.child.length; ++i) {
      Response childResponse = child[i].tick();
      if (childResponse == Response.SUCCESS) {
        success += 1;
      } else if (childResponse == Response.FAILURE) {
        failure += 1;
      }
    }
    if (success >= this.threshold) {
      this.halt(); return Response.SUCCESS;
    } else if (failure > this.child.length - this.threshold) {
      this.halt(); return Response.FAILURE;
    } else {
      this.state = State.RUNNING; return Response.RUNNING;
    }
  }

}
```

Listing 7: Concrete class `ReactiveParallel`.

**Reactive Parallel** A `ReactiveParallel` node requires the definition of a *success threshold*. Informally, ticking such a node amounts to ticking all of its $n$ children; given a success threshold $k$ such that $1 \leq k \leq n$ there are three cases: $(i)$ return `SUCCESS`, if at least $k$ children return `SUCCESS`, $(ii)$ return `FAILURE`, if at least $n - k$ children return `FAILURE`, and $(iii)$ return `RUNNING` otherwise. The class is formalized in Listing 7, where `threshold` is an attribute of the class — and an additional parameter to the constructor — and `tick()` defines the control logic. Notice that whenever the (un)success threshold is met the node calls `halt()` on itself to halt any running child and set itself in the idle state. Whenever the (un)success threshold is not met, the node state is `RUNNING`.

```java
final class SequenceWithMemory extends ControlNode {

  private int toTick;

  public SequenceWithMemory(SymbolTable blackboard, int nChildren) {
    super(blackboard, nChildren);
    this.toTick = 0;
  }

  public Response tick() {
    for (int j = this.toTick; j < this.child.length; ++j) {
      Response childResponse = this.child[j].tick();
      if (childResponse == Response.RUNNING) {
        this.toTick = j;
        this.state = State.RUNNING;
        return Response.RUNNING;
      } else if (childResponse == Response.FAILURE) {
        this.toTick = 0;
        this.state = State.IDLE;
        return Response.FAILURE;
      }
    }
    this.toTick = 0;
    this.state = State.IDLE;
    return Response.SUCCESS;
  }

}
```

Listing 8: Concrete class `SequenceWithMemory`.

**Sequence with Memory** The concrete class `SequenceWithMemory` overrides the `tick()` method as shown in Listing 8. The tick is propagated to the all the nodes in the `child` array, from `toTick` to `child.length - 1`; initially, the value of `toTick` corresponds to the index of the first child, but if a child returns `RUNNING` then `toTick` is set to the index of that child, the state becomes running and the node returns `RUNNING`. At the next tick, children will be ticked starting from the one that was running. If a child response is `FAILURE` then the memory is reset, the node becomes idle and the node itself returns failure; if all the children return `SUCCESS`, then `SUCCESS` is returned after resetting the memory and setting the state to idle. As in reactive sequences, when a child returns `RUNNING` or `FAILURE`, the tick is not propagated to the next child (if any) and the state of the node is `RUNNING` only if at least one child is running. In this case, upon returning `RUNNING` or `FAILURE` there is no need to halt subsequent children, because the sequence with memory does not re-tick nodes unless at least one returns `FAILURE` or every node returns `SUCCESS`, i.e., it is not possible to have running nodes whose index is higher than the one currently being ticked.

```java
final class FallbackWithMemory extends ControlNode {

  private int toTick;

  public FallbackWithMemory(SymbolTable blackboard, int nChildren) {
    super(blackboard, nChildren);
    this.toTick = 0;
  }

  public Response tick() {
    for (int j = toTick; j < this.child.length; ++j) {
      Response childResponse = this.child[j].tick();
      if (childResponse == Response.RUNNING) {
        this.toTick = j;
        this.state = State.RUNNING;
        return Response.RUNNING;
      } else if (childResponse == Response.SUCCESS) {
        this.toTick = 0;
        this.state = State.IDLE;
        return Response.SUCCESS;
      }
    }
    this.toTick = 0;
    this.state = State.IDLE;
    return Response.FAILURE;
  }

}
```

Listing 9: Concrete class `FallbackWithMemory`.

**Fallback with Memory** The concrete class `FallbackWithMemory` overrides the `tick()` method as shown in Listing 8. As in the reactive version, the implementations of `tick()` is the "dual" of the same method in `SequenceWithMemory`: the tick is propagated to the all the nodes in the `child` array, from `toTick` to `child.length - 1`; initially, the value of `toTick` corresponds to the index of the first child, but if a child returns `RUNNING` then `toTick` is set to the index of that child, the state becomes running and the node returns `RUNNING`. At the next tick, children will be ticked starting from the one that was running. If a child response is `SUCCESS` then the memory is reset, the node becomes idle and the node itself returns success; if all the children return `FAILURE`, then `FAILURE` is returned after resetting the memory and setting the state to idle. As in the reactive fallback, when a child returns `RUNNING` or `SUCCESS`, the tick is not propagated to the next child (if any) and the state of the node is `RUNNING` only if at least one child is running. As in `SequenceWithMemory`, upon returning `RUNNING` or `SUCCESS` there is no need to halt subsequent children.

```java
final class ParallelWithMemory extends ControlNode {

  private int threshold;
  private int success;
  private int failure;
  private boolean[] done;

  public ParallelWithMemory(SymbolTable blackboard, int nChildren,
                            int threshold) {
    super(blackboard, nChildren);
    assert(threshold>=1 && threshold<=nChildren);
    this.threshold = threshold;
    this.success = this.failure = 0;
    this.done = new boolean[nChildren];
    for (int i = 0; i < nChildren; ++i) this.done[i] = false;
  }

  public void halt() {
    super.halt();
    this.success = this.failure = 0;
    for (int i = 0; i < this.child.length; ++i) this.done[i] = false;
  }

  public Response tick() {
    for (int i = 0; i < this.child.length; ++i) {
      if (!this.done[i]) {
        Response childResponse = this.child[i].tick();
        if (childResponse != Response.RUNNING) {
          this.done[i] = true;
          if (childResponse == Response.SUCCESS) {
            this.success += 1;
          } else if (childResponse == Response.FAILURE) {
            this.failure += 1;
          }
        }
      }
    }
    if (this.success >= this.threshold) {
      this.halt();
      return Response.SUCCESS;
    } else if (this.failure > this.child.length - this.threshold) {
      this.halt();
      return Response.FAILURE;
    } else {
      this.state = State.RUNNING;
      return Response.RUNNING;
    }
  }

}
```

Listing 10: Concrete class `ParallelWithMemory`.

**Parallel with Memory** The definition of the concrete class `ParallelWithMemory` is shown in Listing 10. This node requires the following attributes to be defined:

- a success threshold (attribute `threshold`) as in the corresponding reactive variant;
- `failure` and `success` counters, as the node "remembers" them from one call to the other;
- a flag for each child (attribute `done`) to remind nodes that were ticked and could still be running.

The logic defined by `tick()` is similar to the reactive variant, with the difference that (un)success count is kept between subsequent calls and only nodes that answer `RUNNING` are ticked repeatedly (their `done` flag is `false`), whereas those answering `SUCCESS` or `FAILURE` are not ticked anymore, but their result concurs to update success and failure counters, respectively. At each tick, if either threshold is met, the node is halted and the corresponding value is returned. Notice that `halt()` is overridden in this case in order to reset the success and failure count as well as the completion flag for each child. Whenever the (un)success threshold is not met, the node state is `RUNNING`.

```java
final class Switch extends ControlNode {

  static final int NO_TICK = -1;

  private String switchKey;
  private int previousTick;

  public Switch(SymbolTable blackboard, int nChildren, String key) {
    super(blackboard, nChildren);
    this.switchKey = key;
    this.previousTick = Switch.NO_TICK;
  }

  public Response tick() {
    int nextTick = blackboard.getValue(switchKey);
    assert(nextTick>=0 && nextTick < this.child.length);
    if (nextTick != this.previousTick) {
      if (this.previousTick != Switch.NO_TICK &&
          this.child[this.previousTick].getState() != State.IDLE) {
        this.child[this.previousTick].halt();
      }
      this.previousTick = nextTick;
    }
    Response childResponse = this.child[nextTick].tick();
    if (childResponse == Response.RUNNING) {
      this.state = State.RUNNING;
      return Response.RUNNING;
    } else {
      this.state = State.IDLE;
      this.previousTick = Switch.NO_TICK;
      return childResponse;
    }
  }

}
```

Listing 11: Concrete class `Switch`.

**Switch** The `Swtich` concrete class is defined in Listing 11. In this case, we need two additional attributes:

- `switchKey` records the key in the blackboard that stores the index of the child whereon `tick()` should be called;
- `previousTick` records the node that was ticked ("switched to") last in order to avoid leaving a running child when the control is routed to another child.

The control logic of `tick()` is about fetching a value from the blackboard to know which node should be ticked next. Whenever the child to be ticked next (index `nextTick`) is different with respect to the one ticked previously (index `previousTick`) the new value is saved and a check for potential running nodes is done. If a previously ticked node whose state is not idle is found, then that node is halted. Finally, the chosen child is ticked: if the response is `RUNNING`, then the switch node itself becomes running and `RUNNING` is returned; otherwise, the state becomes idle, the `previousTick` memory is reset and the return value is the same of the child — either `FAILURE` or `SUCCESS`.

```
abstract class DecoratorNode extends Node {

  protected Node child;

  public DecoratorNode(SymbolTable blackboard, Node child) {
    super(blackboard);
    assert(child != null);
    this.child = child;
  }

  public void halt() {
    if (this.child.getState() != State.IDLE) this.child.halt();
    this.state = State.IDLE;
    return;
  }

}
```

Listing 12: Abstract class `DecoratorNode`.

## 2.3 Decorator Nodes

The purpose of a decorator node is to change the functionality of the control or execution node that it wraps. Specifically, it manipulates the return status of its child according to a user-defined rule and also selectively ticks the child according to some predefined rule.The abstract class `DecoratorNode` is defined in Listing 12. We consider three subclasses of `DecoratorNode` whose behavior is described informally as follows:

- The *inverter* node, as the name implies, changes the return value of its child from `SUCCESS` to `FAILURE` and vice versa.
- The *force* node turns a `FAILURE` result from its child into `SUCCESS` or, on the contrary, a `SUCCESS` result into `FAILURE`.
- The *retry until* node, keeps ticking its child until it returns either `SUCCESS` or `FAILURE`.

The definition of the inverter node is presented in Listing 13, the force node is defined in Listing 14, and the retry until node is defined in Listing 15.

```java
final class Inverter extends DecoratorNode {

  public Inverter(SymbolTable blackboard, Node child) {
    super(blackboard, child);
  }

  public Response tick() {
    Response childResponse = this.child.tick();
    if (childResponse == Response.SUCCESS) {
      this.state = State.IDLE;
      return Response.FAILURE;
    } else if (childResponse == Response.FAILURE) {
      this.state = State.IDLE;
      return Response.SUCCESS;
    } else {
      this.state = State.RUNNING;
      return Response.RUNNING;
    }
  }

}
```

Listing 13: Concrete classe `Inverter`.

```java
final class Force extends DecoratorNode {

  private Response what;

  public Force(SymbolTable blackboard, Node child, Response what) {
    super(blackboard, child);
    assert(what == Response.SUCCESS || what == Response.FAILURE);
    this.what = what;
  }

  public Response tick() {
    Response childResponse = this.child.tick();
    if (childResponse == Response.RUNNING) {
      this.state = State.RUNNING;
      return Response.RUNNING;
    } else {
      this.state = State.IDLE;
      return this.what;
    }
  }

}
```

Listing 14: Concrete classe `Force`.

```java
final class RetryUntil extends DecoratorNode {

  private Response what;

  public RetryUntil(SymbolTable blackboard, Node child, Response what) {
    super(blackboard, child);
    assert(what == Response.SUCCESS || what == Response.FAILURE);
    this.what = what;
  }

  public Response tick() {
    Response childResponse = this.child.tick();
    if (childResponse == this.what) {
      this.state = State.IDLE;
      return what;
    } else {
      this.state = State.RUNNING;
      return Response.RUNNING;
    }
  }

}
```

Listing 15: Concrete classe `RetryUntil`.

# 3 Definition and Graphical Syntax

Given an instance $S$ of a concrete subclass of *SymbolTable*, a *well formed behavior tree* (WFBT) is defined recursively as follows:

- Any instance of an `Action` or `Condition` concrete subclass that has $S$ as a `blackboard` attribute is a WFBT.
- Given $T_1, \ldots T_n$ WFBTs with $n > 1$, a control node whose children are $T_1, \ldots T_n$ and whose `blackboard` attribute is $S$ is also a WFBT.
- If $T$ is a WFBT, then a decorator node whose child is $T$ and whose `blackboard` attribute is $S$ is also a WFBT.

In the remainder of this paper, when we refer to a BT, unless explicitly noted, we mean a WFBT with a (unique) symbol table $S$. BTs can be represented graphically using a composition of the symbols defined in Table 2 where, for each instance of the classes presented in Section 2, we provide a corresponding graphical notation.

As an example, in Figure 2 we show the graphical representation of a complete BT. The corresponding code-based representation is based on the following assumptions:

- `BatteryLevel`, `isPoiDone`, and `VisitorsFollowing` are concrete subclasses of `Condition`.

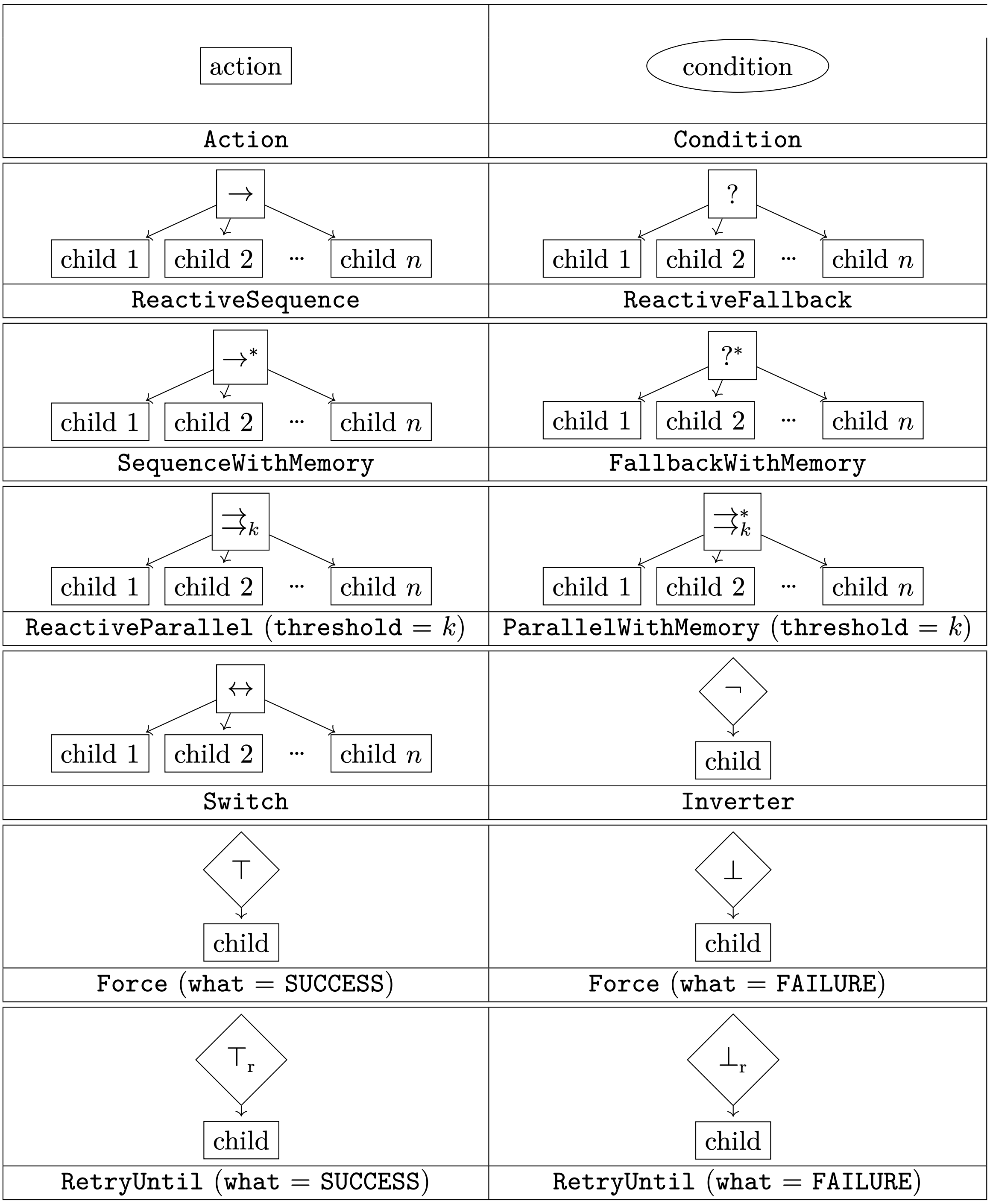

Table 2: Graphical syntax of BT nodes.

- `Alarm`, `SetPoi`, `Reset`, `Wait`, `GoToPoi`, `SetPoiDone` are concrete subclasses of `Action`;
- `Context` is a concrete subclass of `SymbolTable`.

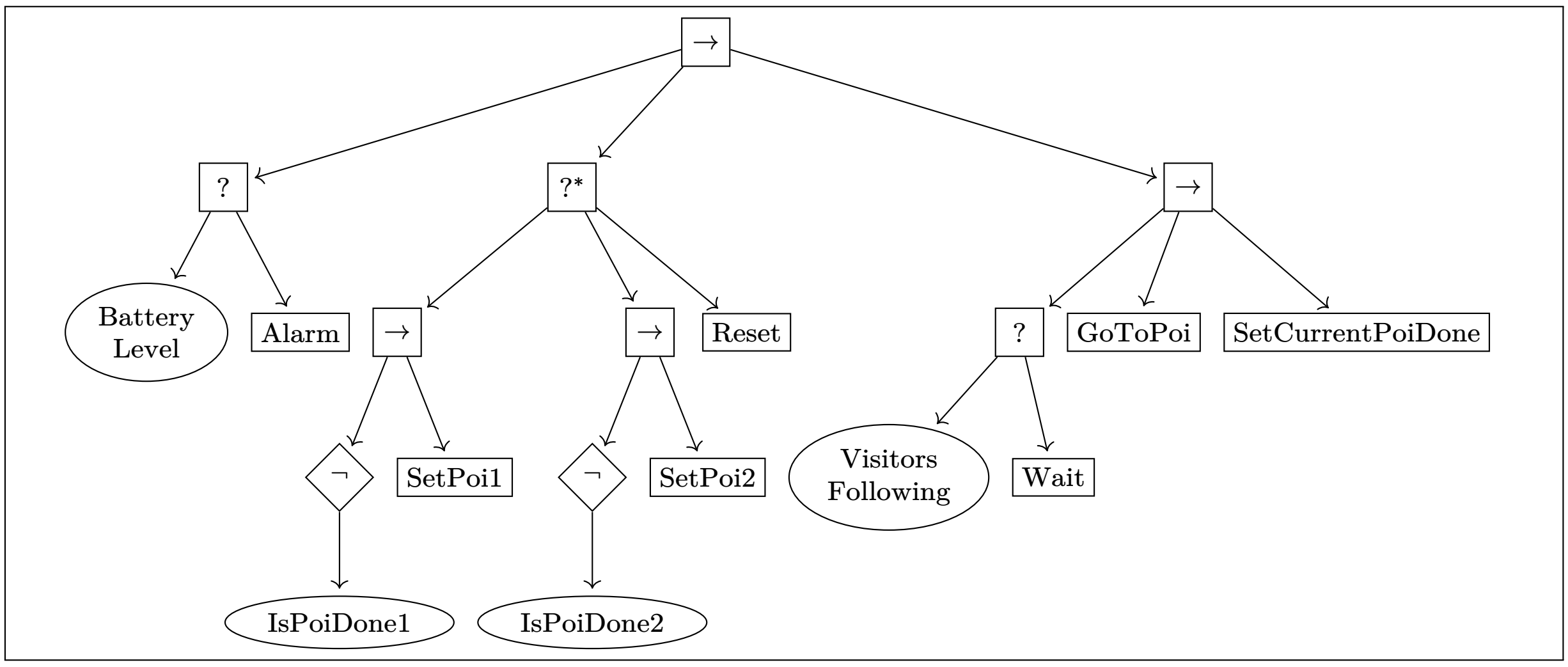

Figure 2

In Listing 16 the definition of the BT corresponding to Figure 2 is presented. In the code, a further assumption is that `"PoI 1"` and `"PoI 2"` are labels with which `setPoi` and `IsPoiDone` objects can fetch the corresponding "Point of interest" data from the symbol table, and that the constructor of these classes accept such key as an additional parameter.

```
SymbolTable context = new Context();

// Creating the "battery" subtree
Node batteryLevel = new BatteryLevel(context);
Node alarm = new Alarm(context);
ReactiveFallback batteryTree = new ReactiveFallback(context, 2);
batteryTree.addChild(batteryLevel, 0);
batteryTree.addChild(alarm, 1);

// Creating the "scheduler" subtree
Node isPoiDone = new IsPoiDone(context, "PoI 1");
Node setPoi = new SetPoi(context, "PoI 1");
ReactiveSequence r1 = new ReactiveSequence(context, 2);
r1.addChild(new Inverter(context, isPoiDone), 0);
r1.addChild(setPoi, 1);

isPoiDone = new IsPoiDone(context, "PoI 2");
setPoi = new SetPoi(context, "PoI 2");
ReactiveSequence r2 = new ReactiveSequence(context, 2);
r2.addChild(new Inverter(context, isPoiDone), 0);
r2.addChild(setPoi, 1);

Node reset = new Reset(context);
FallbackWithMemory schedulerTree = new FallbackWithMemory(context, 3);
schedulerTree.addChild(r1, 0);
schedulerTree.addChild(r2, 1);
```

```java
schedulerTree.addChild(reset, 2);

// Creating the "navigation" subtree
Node vFollowing = new VisitorsFollowing(context);
Node wait = new Wait(context);
ReactiveFallback f1 = new ReactiveFallback(context, 2);
f1.addChild(vFollowing, 0);
f1.addChild(wait, 1);
Node goToPoi = new GoToPoi(context);
Node setPoiDone = new SetPoiDone(context);
ReactiveSequence navTree = new ReactiveSequence(context, 3);
navTree.addChild(f1, 0);
navTree.addChild(goToPoi, 1);
navTree.addChild(setPoiDone, 2);

// Creating the overall tree
ReactiveSequence tree = new ReactiveSequence(context, 3);
tree.addChild(batteryTree, 0);
tree.addChild(schedulerTree, 1);
tree.addChild(navTree, 2);
```

Listing 16: Code for the BT in Figure 2.

# 4 Conclusions

This paper proposes a precise definition of execution semantics for Behavior Trees in terms of Java code. There are various topics and issues concerning the definition of Behavior Trees and their semantics that this document does not touch upon, as there is no consensus on how to treat them. We present some of the most notable ones in the following.

## 4.1 Halt Semantics for Reactive Control Nodes

Reactive nodes try to have at most one running child at a time by stopping all siblings of a child that returns `RUNNING` (given the execution semantics for reactive nodes, only the siblings following the child that is ticked can be running, while those preceding it must be idle). Though, this means that, between the time the child is ticked and the time all its running siblings are halted, potentially multiple actions are running concurrently.

Of course, the implementor of the Action nodes could account for such behavior on a case-by-case basis. Though, since the potential for unintended side-effects in a large BT and/or in a complex system is elevated, it would be best if the semantics of the BT provided built-in safeguards.

## 4.2 Side-effects for Leaves

Executing a Condition leaf should not have side-effects, i.e., causing changes in the environment. It is not possible for the BT to actually guarantee that this is the case in implementations, but the present specification is meant to apply under such assumption. Similarly, halting an idle Action is also meant to not have side-effects, but in this case the specification avoids the issue entirely by checking if a node or leaf is running before sending it the `halt()` signal.